\documentclass[10pt,twocolumn,letterpaper]{article}

\usepackage{iccv}
\usepackage{times}
\usepackage{epsfig}
\usepackage{graphicx}
\usepackage{amsmath}
\usepackage{amssymb}

\usepackage{enumitem}
\usepackage{pifont}
\usepackage{bbding}
\usepackage{booktabs}
\usepackage{caption}

\usepackage{amsmath,amsfonts,bm}

\def\eqref#1{equation~\ref{#1}}

\def\1{\bm{1}}

\DeclareMathAlphabet{\mathsfit}{\encodingdefault}{\sfdefault}{m}{sl}
\SetMathAlphabet{\mathsfit}{bold}{\encodingdefault}{\sfdefault}{bx}{n}

\newcommand{\Ls}{\mathcal{L}}
\newcommand{\R}{\mathbb{R}}

\DeclareMathOperator*{\argmax}{arg\,max}

\newlength\savewidth\newcommand\shline{\noalign{\global\savewidth\arrayrulewidth
  \global\arrayrulewidth 1pt}\hline\noalign{\global\arrayrulewidth\savewidth}}
\usepackage[pagebackref=true,breaklinks=true,letterpaper=true,colorlinks,bookmarks=false]{hyperref}
\iccvfinalcopy 

\def\iccvPaperID{1676} 

\ificcvfinal\fi

\begin{document}

\title{Associating Spatially-Consistent Grouping with Text-supervised \\ Semantic Segmentation}
\author{Yabo Zhang \thanks{The first author performed this work as an intern at ByteDance.} $^{, 1}$ \quad
Zihao Wang$^{2}$ \quad
Jun Hao Liew$^{2}$ \quad
Jingjia Huang$^{2}$ \quad
Manyu Zhu$^{2}$ \quad \\ 
Jiashi Feng$^{2}$ \quad
Wangmeng Zuo$^{1}$\textsuperscript{(\Envelope)} \\
$^1$Harbin Institute of Technology \quad $^2$ ByteDance \\
\small
\texttt{
\{wangzihao.vision,junhao.liew,huangjingjia,zhumanyu,jshfeng\}@bytedance.com}\\
\small
\texttt{
hitzhangyabo2017@gmail.com, wmzuo@hit.edu.cn}
}

\maketitle
\ificcvfinal\thispagestyle{empty}\fi

\begin{abstract}

In this work, we investigate performing semantic segmentation solely through the training on image-sentence pairs.
Due to the lack of dense annotations, existing text-supervised methods can only learn to group an image into semantic regions via pixel-insensitive feedback. 
As a result, their grouped results are coarse and often contain small spurious regions, limiting the upper-bound performance of segmentation.
%
%
On the other hand, we observe that grouped results from self-supervised models are more semantically consistent and break the bottleneck of existing methods.
Motivated by this, we propose to associate self-supervised spatially-consistent grouping with text-supervised semantic segmentation.
Considering the part-like grouped results, we further adapt a text-supervised model from image-level to region-level recognition with two core designs.
%
%
First, we encourage fine-grained alignment with a one-way noun-region contrastive loss, which reduces the mismatched noun-region pairs.
Second, we adopt a contextually aware masking strategy to enable simultaneous recognition of all grouped regions.
Coupled with spatially-consistent grouping and region-adapted recognition, our method achieves 59.2\% mIoU and 32.4\% mIoU on Pascal VOC and Pascal Context benchmarks, significantly surpassing the state-of-the-art methods.
\end{abstract}

\section{Introduction}
\label{sec:intro}
Semantic segmentation, aiming to divide an image into several regions with human-recognized labels, has made significant strides since the deep learning era~\cite{zhou2019semantic,caesar2018coco,mottaghi2014role,everingham2010pascal}.
However, the superior performance of prior works~\cite{minaee2021image,long2015fully} highly depends on pixel-wise category annotations, whose dense labels are extremely labor-intensive~\cite{bearman2016s}.
To mitigate the demands of human effort, it is encouraging to collect a profusion of image-sentence pairs from the Internet~\cite{radford2021learning} and employ text supervision to facilitate more scalable training for semantic segmentation.

%

\begin{table}[t]
   \centering
   \caption{\textbf{Upper bound segmentation performance} on Pascal VOC 2012.
   }
   \resizebox{0.7\linewidth}{!}{
   \begin{tabular}{l|c}
      Model &Upper Bound (mIoU)\\
      \shline
      MaskCLIP~\cite{zhou2022extract} &{45.3}\\
      GroupViT~\cite{xu2022groupvit}  &{66.3}\\
      Ours  &\textbf{75.2}\\
   \end{tabular}
   }
    \label{tab:upper_bound}
    \vspace{-1em}
 \end{table}
\begin{figure}[tb]
  \centering
   \includegraphics[width=0.95\linewidth]{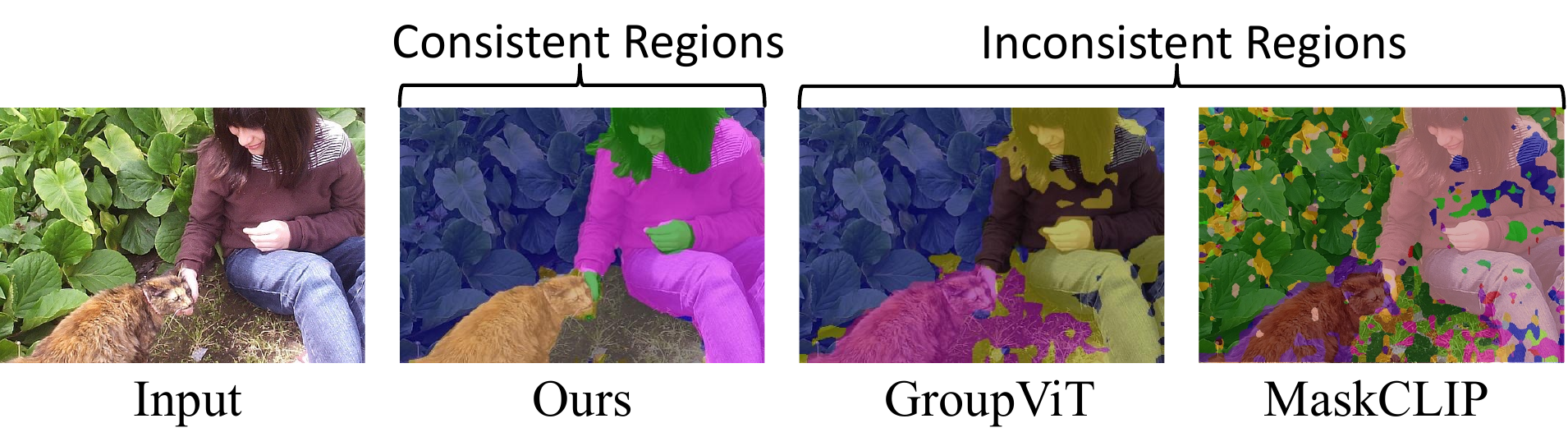}
   \caption{\textbf{Comparison between consistent and inconsistent grouped regions} on text-supervised semantic segmentation.
   GroupViT and MaskCLIP erroneously group pixels from distinct objects together, which cannot be corrected even with optimal recognition.
   On the contrary, our approach produces more consistently grouped regions with self-supervised models.
   We analyze their upper bound performance by optimally assigning ground truth label to the grouped regions, and conclude that the bottleneck of existing text-supervised semantic segmentation methods lies in the spatially inconsistent grouping.
   }
   \label{fig:intro}
\end{figure}

Recent works~\cite{xu2022groupvit,zhou2022extract,ranasinghe2022perceptual} present promising results on text-supervised semantic segmentation following vision-language pre-training methods~\cite{radford2021learning,yao2021filip,jia2021scaling}.
By minimizing the image-sentence contrastive loss~\cite{radford2021learning}, these approaches learn to group the visual concepts into distinct regions and recognize them into semantic categories.

%
The effectiveness of recognition ability has been well investigated by prior works ~\cite{radford2021learning,yao2021filip,jia2021scaling}, as demonstrated by, \eg, superior performance on zero-shot ImageNet~\cite{deng2009imagenet} classification and other downstream tasks.
Nonetheless, we observe that text-supervised segmentation models often lead to spatially inconsistent grouping, as they only learn to group via coarse and pixel-insensitive recognition feedback. 
Fig~\ref{fig:intro} shows several examples of coarse and spurious grouped regions.
Without elaborate post-processing, it is infeasible to properly separate wrongly-grouped pixels during the recognition process, which limits their upper bound segmentation performance.
Compared to the above text-supervised models,  self-supervised vision models show some emerging properties on grouping pixels into spatially-consistent regions~\cite{caron2021emerging,he2022masked,he2020momentum,chen2020simple}.
Self-supervised vision models acquire the ability to encode semantically-close pixels into similar features by encouraging the representation of different augmented views being consistent~\cite{caron2021emerging,he2020momentum,chen2020simple}.
As a result, despite the absence of human-defined supervision, their features exhibit significant spatial consistency with pixels.
They can be directly used for delineating parts or objects with the simple clustering process~\cite{wang2022tokencut,van2022discovering,hamilton2022unsupervised}.
%
%
In Fig.~\ref{fig:intro} and Tab.~\ref{tab:upper_bound}, we present the comparisons between the self-supervised and text-supervised grouped results, and demonstrate that the former is more semantically coherent and attains a higher upper bound on semantic segmentation.

Based on the analysis above, we exploit the spatially consistent grouping capabilities of self-supervised vision models to break the bottleneck of text-supervised segmentation.
In specific, we directly cluster an image into $M$ class-agnostic regions with its self-supervised features~\cite{caron2021emerging,amir2021deep,hamilton2022unsupervised}.
Since images are sometimes over-segmented into part-like regions (\eg, ``hand'' in Fig.~\ref{fig:intro}), we further propose to adapt a text-supervised model~\cite{radford2021learning} from image-level to region-level recognition by \textit{two core designs}.
%
%
First, we introduce a context-aware masking strategy to simultaneously encode all grouped regions.
In each intermediate layer, all patch tokens interact with each other and are organized into contextually aware features.
When encoding grouped regions, their respective class
tokens concurrently query the well-organized patch tokens within their masks.
%
%
Second, we propose a one-way fine-grained contrastive loss to encourage noun-region alignment, while minimizing potential misaligned noun-region pairs.
In general, object-grained information in a sentence is sparser than in an image (\eg, in Fig.~\ref{fig:image_caption}), thus we only match each noun to its closest region and exclude the mismatched regions.
%
%

To avoid potential degradation of the open-vocabulary recognition ability~\cite{ilharco2022patching}, we only fine-tune several learnable tokens for a better trade-off between region-level adaption and open-vocabulary preservation.
It also brings another benefit, \ie, less training cost in region-level adaption.
After fine-tuning on Conceptual Caption, our method achieves 59.2 \% mIoU and 32.4 \% mIoU on Pascal VOC~\cite{everingham2010pascal} and Pascal Context~\cite{mottaghi2014role}, significantly outperforming previous approaches.

In summary, our contributions are as follows:
\vspace{-2mm}
\begin{itemize}[itemsep=0pt, parsep=0pt, leftmargin=0pt]
    \item We propose to associate spatially-consistent grouping with text-supervised semantic segmentation for a higher upper-bound segmentation performance.
    \item We adapt a text-supervised model for region-level recognition with a one-way noun-region alignment loss and a context-aware masking strategy.
    \item We conduct extensive experiments and achieve state-of-the-art performance on several popular benchmarks.
\end{itemize}

\section{Related Work}
\paragraph{Self-supervised and Text-supervised Pre-training.}
Self-supervised and text-supervised pre-trained models provide two complementary abilities for semantic segmentation, \ie, grouping and recognition.
Self-supervised learning, also known as unsupervised learning, aims to learn good visual representations of images without any human-defined labels~\cite{caron2021emerging,he2022masked,he2020momentum,chen2020simple}.
Its pretext task~\cite{he2020momentum} usually involves modelling similarity between different augmentation views of an image.
Particularly, DINO~\cite{caron2021emerging} constructs local and global views of an image and encourages ``local-to-global'' correspondences, and thus its visual features contain spatially-consistent and explicit information for visual grouping.
After distilling process, ~\cite{hamilton2022unsupervised,van2022discovering,wang2022tokencut} explore to transfer this information to unsupervised semantic segmentation~\cite{hamilton2022unsupervised,van2022discovering} or object discovery~\cite{wang2022tokencut} tasks with appealing results.

Text-supervised pre-training methods~\cite{radford2021learning,jia2021scaling,yao2021filip} focus on the vision-language alignment by modelling similarity of positive image-sentence pairs (or dissimilarity of negative pairs).
Training on millions of image-sentence pairs from Internet, text-supervised models have shown tremendous success on zero-shot image classification on ImageNet~\cite{deng2009imagenet}.
However, their performance is still limited on masked or cropped images recognition due to the granularity-gap between pre-training and inference.
Previous works~\cite{zhong2022regionclip,liang2022open} attempt to reduce this gap by fine-tuning, yet depending on fine-grained annotations (\eg, bounding boxes or masks).
Differently, our method adopts a optimal assignment mechanism that dynamically aligns nouns and regions during fine-tuning.
%
\begin{figure*}[hptb]
  \centering
   \includegraphics[width=0.8\linewidth]{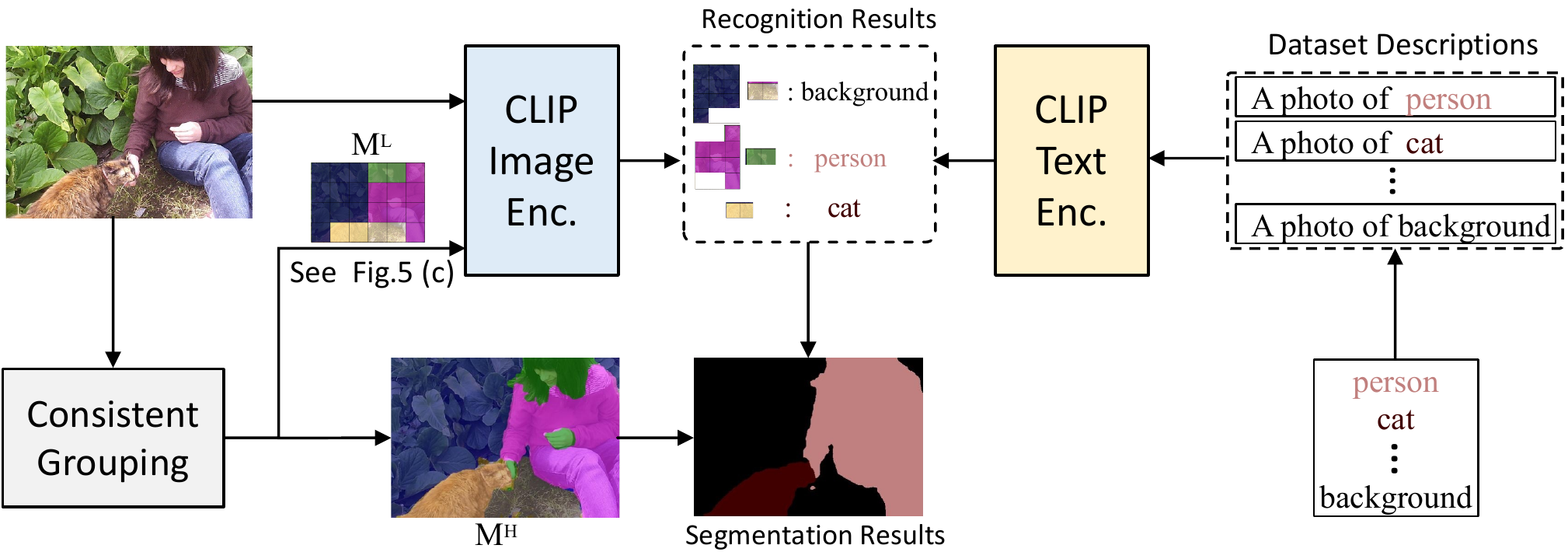}
   \caption{\textbf{Inference pipeline of our method.}
   Firstly, our method parses an image into consistently grouped regions by clustering its self-supervised features, including low-resolution masks $\mathbf{M^L}$ and counterpart high-resolution masks $\mathbf{M^H}$.
   Secondly, CLIP image encoder encodes these regions into region embeddings $\{\mathbf{r_i}\}_{i=1}^M$.
   In parallel, CLIP text encoder encodes dataset categories into text embeddings and uses them to recognize grouped regions.
   Finally, our method combines recognized labels and $\mathbf{M^H}$ into final semantic segmentation masks.
   }
   \label{fig:framework}
\end{figure*}
\paragraph{Text-supervised Semantic Segmentation.}
Previous works~\cite{ghiasi2022scaling,liang2022open} have verified the effectiveness of text supervision on open-vocabulary semantic segmentation.
But semantic segmentation with text-only supervision~\cite{xu2022groupvit,zhou2022extract,ranasinghe2022perceptual} has just been introduced recently and not be well explored.
The difficulty lies in grouping an input image into spatially-consistent regions (or masks), where fully-supervised methods learn it from ground truth masks. 
GroupViT~\cite{xu2022groupvit} is the first work for text-supervised semantic segmentation, and it optimizes the learning of grouping via top-down feedback from image recognition.
MaskCLIP~\cite{zhou2022extract} and CLIPpy~\cite{ranasinghe2022perceptual} leverage the power of pre-trained vision-language models, but perform grouping in a implicit way.
They both consider dataset categories as group queries and group pixels based on query-pixel similarities.

Despite considerable performance on segmentation, prior works still struggle with producing consistently grouped results, which contain apparently small spurious regions.
The main reason is the lack of constraints on spatial redundancy information in images.
Feedback from image recognition mainly benefits coarse object-like grouping, because less or extra redundant pixels have little impact on recognition.
%

\section{Proposed Method}
\begin{figure*}[tb]
  \centering
  \includegraphics[width=0.9\linewidth]{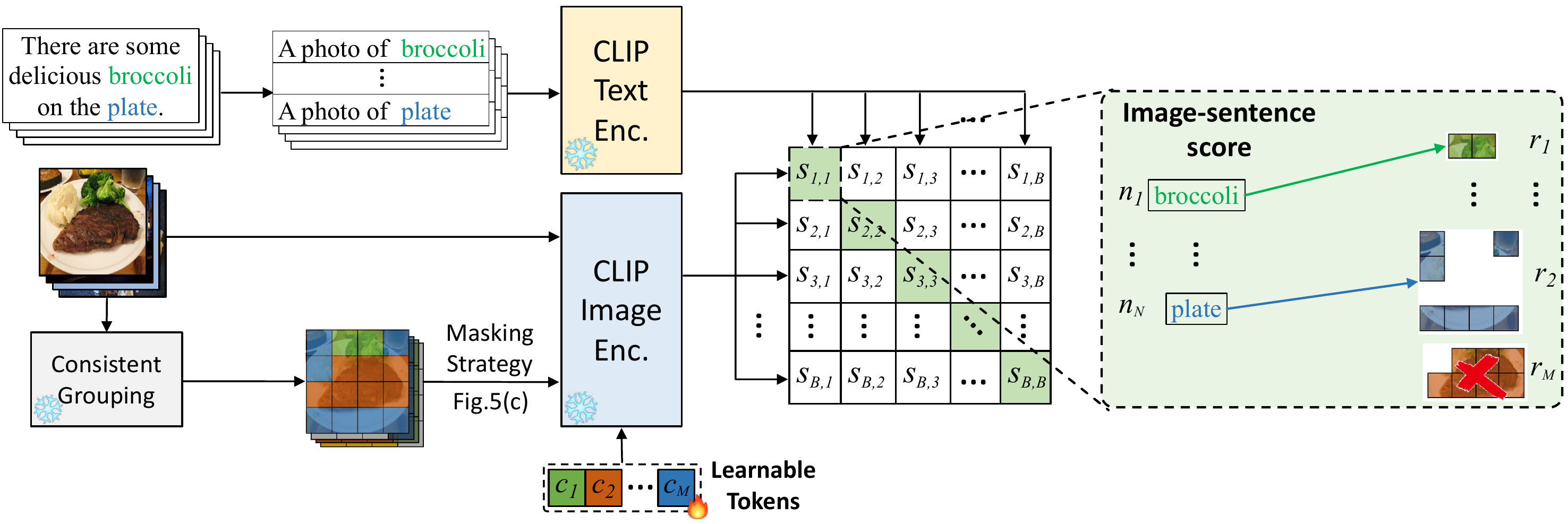}
   \caption{\textbf{Fine-tuning process of region-level adaption.} 
   We introduce a one-way noun-region contrastive loss for effectively fine-grained alignment, while reducing the potentially misaligned noun-region pairs. 
   For a sentence, we fill its nouns into several descriptions and extract their noun embeddings with a CLIP text encoder.
   For an image, we cluster it into consistently grouped regions and recognize the regions with an efficient masking strategy (Fig.~\ref{fig:mask} (c)).
   Due to sparser object-grained information in a sentence (Fig.~\ref{fig:image_caption}), we only match each noun to its closest region and ignore the unmatched regions.
   }
   \label{fig:enhancement}
\end{figure*}

\begin{figure*}[t!]
  \centering
   \includegraphics[width=0.95\linewidth]{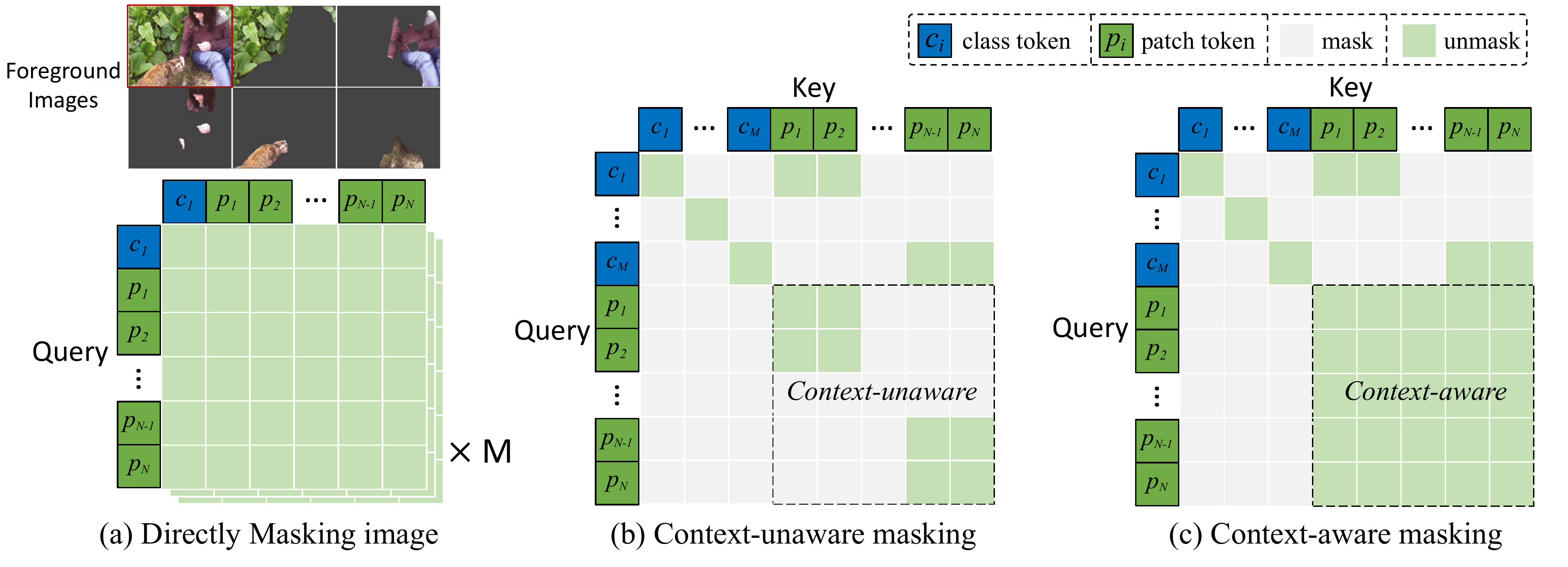}
   \caption{\textbf{Three masking strategies.}
   Strategy (a) directly masks an image to obtain foreground images for each grouped region, and strategy (b) and (c) operate masks on patch tokens.
   In terms of efficiency, strategy (a) requires $M$ forward passes to encode all grouped regions, while strategy (b) and (c) only need one forward pass.
   As for effectiveness, strategy (c) connects all patch tokens like encoding an entire image, thereby encoding more context-aware information than (a) and (b).
   Strategy (c) is adopted in our method.
   }
   \label{fig:mask}
\end{figure*}

In this section, we analyze the bottleneck that limits the upper bound of previous methods, and propose to break it by exploiting consistent features from self-supervised models (Sec.~\ref{sec:preliminary}).
Since self-supervised models may cluster an image into unrecognizable part-like regions, we further adapt a text-supervised model for region-level recognition by two core designs (Sec.~\ref{sec:adaption}), including a one-way noun-region alignment loss (Fig.~\ref{fig:enhancement}) and a context-aware masking strategy (Fig.~\ref{fig:mask})

%
%


\subsection{Preliminary}
\label{sec:preliminary}
Text-supervised semantic segmentation aims to parse an image into several recognized regions by training on image-sentence pairs only.
Prior fully-supervised works learn to group visual regions using dense mask annotations, whereas text-supervised methods~\cite{xu2022groupvit,ranasinghe2022perceptual,zhou2022extract} rely on \textit{pixel-insensitive} recognition feedback only.
Considering that the pixels in images have heavy spatial redundancy~\cite{he2022masked}, they are encoded inconsistently, resulting in coarse and spurious groupings of regions in Fig.~\ref{fig:intro}.
These incorrectly grouped pixels cannot be well separated without elaborate post-processing, thus limiting the upper bound segmentation performance.

In contrast, self-supervised vision models~\cite{caron2021emerging,he2020momentum,he2022masked} learn to encode pixels by keeping the representation of different augmented views being consistent.
Although there is no labor-intensive dense supervision, the encoded features still exhibit remarkable spatial consistency with pixels, and are utilized for grouping pixels into consistent regions.
In Fig.~\ref{fig:intro}, we observe that using consistent grouped results from self-supervised models significantly improves the upper bound of text-supervised semantic segmentation.
%

%
%
%
%
%
%
According to the above analysis, we propose to exploit consistent features from self-supervised vision models in our novel pipeline.
As shown in Fig.~\ref{fig:framework}, our method groups an image $\mathbf{I} \in \R^{H\times W \times C}$ into $M$ consistent regions by clustering its self-supervised features~\cite{caron2021emerging}.
Then, our method uses a CLIP to recognize all grouped regions in a forward pass.
The specific processes are as follows:
%
\paragraph{Consistent Grouping.}
Self-supervised vision methods~\cite{caron2021emerging,he2022masked,he2020momentum} learn to encode pixels into semantic features by keeping different augmented views consistent, and the consistency can be further enhanced with the self-distilling process~\cite{hamilton2022unsupervised}.
Therefore, using the encoded features from self-supervised models~\cite{caron2021emerging,hamilton2022unsupervised}, our method performs consistent grouping via plain clustering algorithms~\cite{dhanachandra2015image} only.
%
%
Formally, given an image $\mathbf{I}$, we encode it into compact visual features $\mathbf{F^L} \in \R ^{H/P \times W/P \times D}$, following by clustering into $M$ low-resolution masks $\mathbf{M^L} = \{\textbf M_i^L\}_{i=1}^M$.
Meanwhile, we upsample visual features into $\mathbf{F^H} \in \R ^{H \times W \times D}$ to obtain high-resolution counterpart $\mathbf{M^H} = \{\textbf M_i^H\}_{i=1}^M$.

\paragraph{Region Recognition.}
To recognize the grouped regions in a zero-shot manner, we employ a text-supervised model (\ie, CLIP~\cite{radford2021learning}) pre-trained large-scale image-sentence pairs.
%
%
In Fig.~\ref{fig:framework}, when performing semantic segmentation on specific datasets, we obtain the descriptions of $N_D$ categories by prompt engineering, and then encode them into text embeddings $\{\mathbf{t_j}\}_{j=1}^{N_D}$ with a text encoder.
For grouped regions of an input image, we compute the region embeddings $\{\mathbf{r_i}\}_{i=1}^M$ with their corresponding low-resolution masks $\mathbf{M^L}$.
Finally, we assign each region to the category with the highest region-text cosine similarity, followed by combining them with high-resolution masks $\mathbf{M^H}$ to get the final segmentation results.
%

\begin{figure}[tb]
  \centering
  \includegraphics[width=0.9\linewidth]{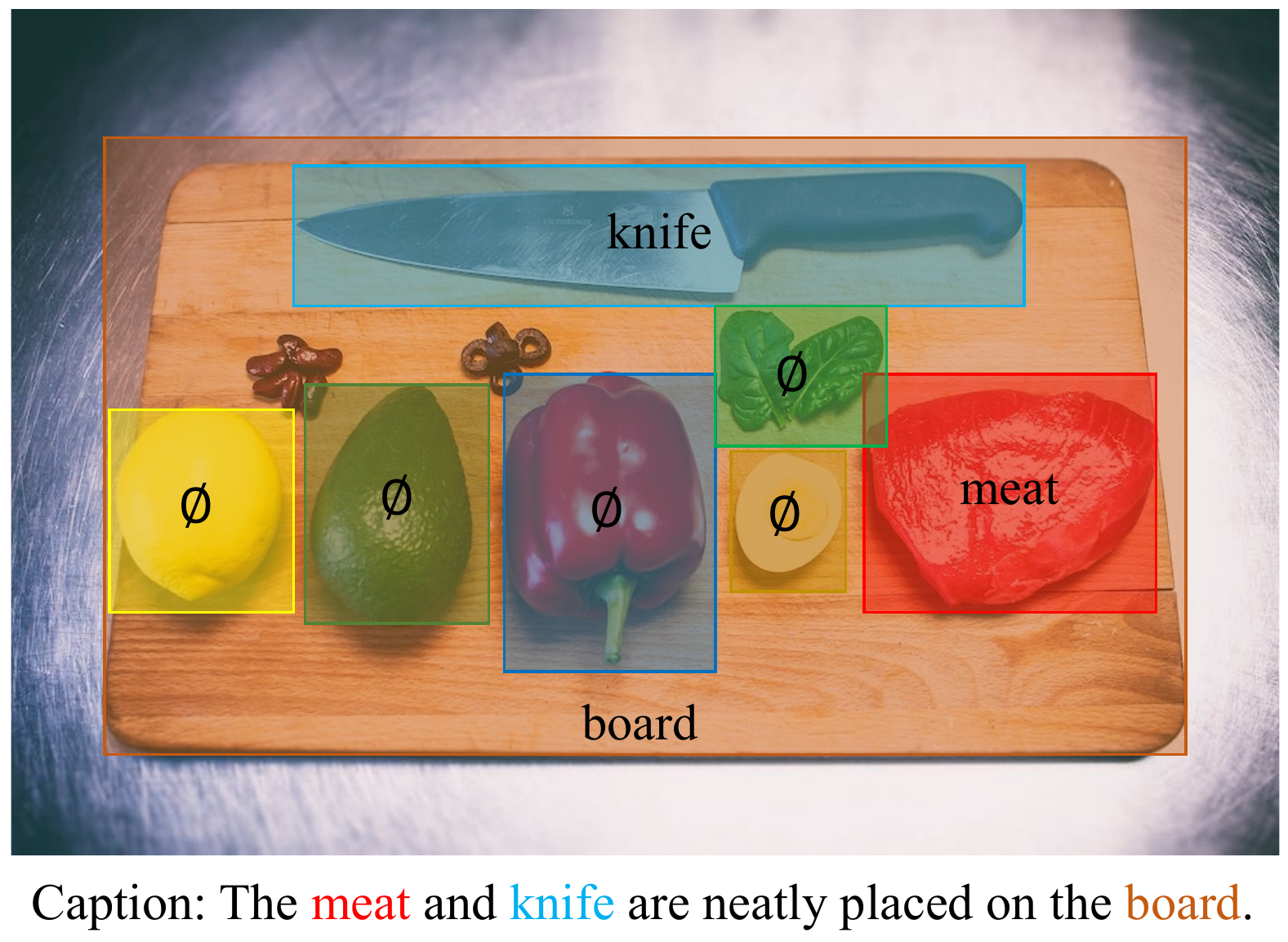}
   \caption{\textbf{Asymmetry object-level information in an image-caption pair.}
   Each noun in the caption is associated with its corresponding object in the image, yet some objects lack their corresponding nouns within the caption.
   }
   \label{fig:image_caption}
   \vspace{-0.5cm}
\end{figure}

\subsection{Region-level Adaption}
\label{sec:adaption}
We notice that directly applying CLIP to recognize grouped regions leads to unsatisfactory results on semantic segmentation.
The main reason is that CLIP is pre-trained to match an image with a sentence, while not considering fine-grained noun-region alignment~\cite{zhong2022regionclip}.
Moreover, the grouped results of self-supervised models usually contain some part-like regions (\eg, ``arm'' in Fig.~\ref{fig:intro}), which further increases the difficulty of recognition.
To \textit{efficiently} address this issue, we propose a masking strategy to perform simultaneous and in-domain recognition, along with a one-way noun-region contrastive loss to encourage fine-grained alignment.

\paragraph{Masking Strategy.}
Grouped regions are delineated by their corresponding masks, and how to represent each region with its mask has a great impact on recognition results.
As shown in Fig.~\ref{fig:mask}, performing masking on pixels or patch tokens is either time-consuming or domain-shifted:
\textbf{(a)} Directly masking unrelated pixels will result in out-of-domain images (\ie, transparent background), and recognizing $M$ grouped regions is very time-consuming (\ie, $M$ forward passes).
\textbf{(b)} Operating masks on patch tokens can simultaneously encode all grouped regions in a single forward pass, but still has a non-negligible domain shift, especially in part-like grouped regions.

It has been observed in~\cite{ding2022open} that adding interactions between patch tokens provides more context information.
To reduce the domain shift between image and region recognition, we reformulate their encoding process as querying from context-aware image features in Fig.~\ref{fig:mask} (c). The only difference is their receptive field.
Concretely, in all intermediate layers of the image encoder, all patch tokens interact with each other and are encoded into well-organized features.
When encoding grouped regions, their corresponding class tokens attend to patch tokens within their masks in parallel.
We compare three masking strategies in Sec.~\ref{sec:ablation} and verify the effectiveness and efficiency of our strategy.
%

%
%
%
%

\paragraph{Noun-Region Alignment.}
Previous works~\cite{zhong2022regionclip,minderer2022simple} introduce a fine-grained contrastive loss in object detection, but it does not perform well when transferring to semantic segmentation.
A potential reason is that they ignore the \textit{asymmetry} object-level information in a pair of image-sentence.
In Fig.~\ref{fig:image_caption}, one can observe that an image can be parsed into multiple regions including background, whereas nouns in a sentence are much sparser without their corresponding regions (\eg, lemon, avocado, capsicum in Fig.~\ref{fig:image_caption}).
%

%
%
%
%

%
%
%
%

%
For more effective noun-region alignment, we propose a novel noun-region contrastive loss via \textit{one-way} assignment in Fig.~\ref{fig:enhancement}.
Specifically, we first use an offline parser to extract $N_T$ nouns from a sentence $T^i$, and encode their prompted descriptions into noun embeddings $\{\mathbf n_l^i\}_{l=1}^{N_T}$.
For grouped regions, we simultaneously encode them into region embeddings $\{\mathbf r_k^j\}_{k=1}^M$ with our masking strategy.
%
%
Next, we assign each noun embedding to its closest region embedding, and compute the average similarity of all assigned noun-region pairs as the image-sentence score:
\begin{equation}
    \label{eq:score}
    s_{i,j} = \frac 1 {N_T} \sum_{l=1}^{N_T} \frac {{\mathbf n_l^i} \cdot \mathbf r_{\sigma_{i,j}(l)}^j} {|{\mathbf n_l^i}||\mathbf r_{\sigma_{i,j}(l)}^j|},
\end{equation}
where $\sigma_{i,j}$ denotes the function assigning nouns $\{\mathbf n_l^i\}_{l=1}^{N_T}$ to regions $\{\mathbf r_k^j\}_{k=1}^M$:
\begin{equation}
    \label{eq:assign}
    \sigma_{i,j}(l)=\argmax_{k} \frac {{\mathbf n_l^i} \cdot \mathbf r_{k}^j} {|{\mathbf n_l^i}||\mathbf r_{k}^j|},
\end{equation}
Finally, our one-way noun-region contrastive loss is defined as:
\begin{equation}
    \label{eq:region_loss}
    \Ls = -\frac 1 B \sum_{i=1}^B \frac {\exp(s_{i,i}/ \tau)} {\sum_{j=1}^B \exp(s_{i,j}/ \tau)} -
    \frac 1 B \sum_{j=1}^B \frac {\exp(s_{j,j} / \tau)} {\sum_{i=1}^B \exp(s_{i,j} / \tau)} .
\end{equation}
where $\tau$ represents a learnable temperature parameter to control the range of logits.
Notably, when CLIP is adapted on a much less dataset, its ability to open-vocabulary recognition will decrease by a large margin~\cite{ilharco2022patching}.
Instead of fine-tuning all parameters of CLIP, we introduce $M$ learnable tokens for each type of region~\cite{jia2022visual}, making these tokens region-aware without severely sacrificing the recognition capability.
Less trainable parameters also bring less training cost.

\begin{table*}[t]
   \centering
   \caption{\textbf{Quantitative comparisons with previous methods on popular benchmarks.}
   All models are pre-trained with image-sentence pairs and tested in a zero-shot manner.
   For pre-training dataset, ImageNet1M (IN1M) are used to train self-supervised models (\eg, DINO), while other datasets are used to train text-supervised models.
   For a fair comparison with other methods, we retrain two small CLIPs, including CLIP-12M on CC12M and CLIP-27M on CC12M+YFCC15M.
   The best and second best for each column are \textbf{bolded} and \underline{underlined}.
   }
   \begin{tabular}{l|c|ccc}
      Model & Pre-training Dataset & Pascal VOC & Pascal Context & COCO \\
\shline      
      MaskCLIP~\cite{zhou2022extract} & 400M &{38.8} &25.5 &{20.6} \\
      GroupViT~\cite{xu2022groupvit} & CC12M &41.1 &18.2 &18.4\\
      GroupViT~\cite{xu2022groupvit} & CC12M+YFCC15M  &52.3 &22.4 &24.3\\
      CLIPpy~\cite{ranasinghe2022perceptual} & IN1M+CC12M &50.8 &- &23.8 \\
      CLIPpy~\cite{ranasinghe2022perceptual} & IN1M+HQITP134M  &52.2 &- &25.5 \\
      \hline
      Ours  & IN1M+CC12M  &55.0 &27.1 &24.9\\
      Ours  & IN1M+27M  &\underline{57.3} &\underline{29.5} &\underline{25.8}\\
      Ours  & IN1M+400M  &\textbf{59.2} &\textbf{32.4} &\textbf{26.8} \\
   \end{tabular}
    \label{tab:main}
 \end{table*}

\section{Experiments}
Our method aims to perform semantic segmentation by training on image-sentence pairs only.
We compare our method to several state-of-the-art models quantitatively (Sec.~\ref{sec:quan}) and qualitatively (Sec.~\ref{sec:qual}).
We conduct extensive ablation studies to verify the effectiveness of our designs (Sec.~\ref{sec:ablation}).
We evaluate all methods with a mean intersection
over union (mIoU) on popular benchmarks.
\subsection{Implementation Details}
\label{sec:implement}
\paragraph{Framework.}
Our method parses an image into 27 regions by clustering self-supervised features from STEGO~\cite{hamilton2022unsupervised}, which is self-distilled based on DINO~\cite{caron2021emerging}.
%
%
Next, our method uses CLIP-400M~\cite{radford2021learning} as the recognition module, which is pre-trained on 400M collected image-text pairs.
For a fair comparison with other competitors, we also pre-train a CLIP-27M on Conceptual Captions 12M~\cite{changpinyo2021conceptual} (CC12M) + Yahoo Flickr Creative Commons 15M~\cite{thomee2016yfcc100m} (YFCC15M) and a CLIP-12M on CC12M.
%
%
\paragraph{Fine-tuning.}
We adapt the recognition module to the region level by fine-tuning on CC12M, with a batch size of $2,048$ and image size of $224\times224$.
For other hyper-parameters, we use AdamW~\cite{loshchilov2017decoupled} with a learning rate of $0.0005$ and a weight decay of $0.01$.
Thanks to less learnable parameters (\ie, $27$ tokens), the fine-tuning process can be done within $5$ epochs, $\sim$ $2$ hours with $8$ A100 GPUs.
\paragraph{Inference.}
We use \textit{mIoU} as the metric and evaluate our method on the validation sets of Pascal VOC~\cite{everingham2010pascal}, Pascal Context~\cite{mottaghi2014role}, and COCO~\cite{lin2014microsoft} benchmarks.
They include $1.5$k, $5$k, and $5$k images respectively.
Pascal VOC consists of $20$ foreground classes and a background class, and Pascal Context contains $59$ indoor and outdoor classes.
The ambiguous background class is labeled when regions are predicted as categories of Pascal Context and not in Pascal VOC.
Following GroupViT~\cite{xu2022groupvit}, for each image in COCO, we combine its instance masks belonging to the same category as the semantic mask, so COCO has $80$ object classes and a background class.
Similarly, the background class of COCO is labeled when regions are predicted as categories of COCO-Stuffs and not in COCO objects.
For each benchmark, we encode its categories into text embeddings by prompt engineering and keep images at their original size.
\subsection{Quantitative Comparisons}
\label{sec:quan}
We compare our method with the state-of-the-art methods on text-supervised semantic segmentation in Tab.~\ref{tab:main}.
All approaches are trained with image-text pairs only and tested on semantic segmentation in a zero-shot manner.
They all use ViT-based architecture with a patch size of $16$.
In $6$th row, even using the least pre-training, our method still outperforms previous methods on most benchmarks, \ie, $2.8\%$ mIoU on Pascal VOC and $3.7\%$ mIoU on Pascal Context.
To fairly compare to methods using more data, in $7$th row of Tab.~\ref{tab:main}, we retrain a CLIP-27M on CC12M + YFCC15M, and observe that it surpasses existing approaches on all datasets in a large margin, consistently raised by $\sim 5\%$ mIoU on Pascal VOC and Context benchmarks.
When choosing CLIP-400M released by OpenAI as the recognition module, our method achieves greatly superior performance, $59.2\%$ mIoU on Pascal VOC.
%
%
%
%

\begin{figure*}[t!]
  \centering
   \includegraphics[width=0.9\linewidth]{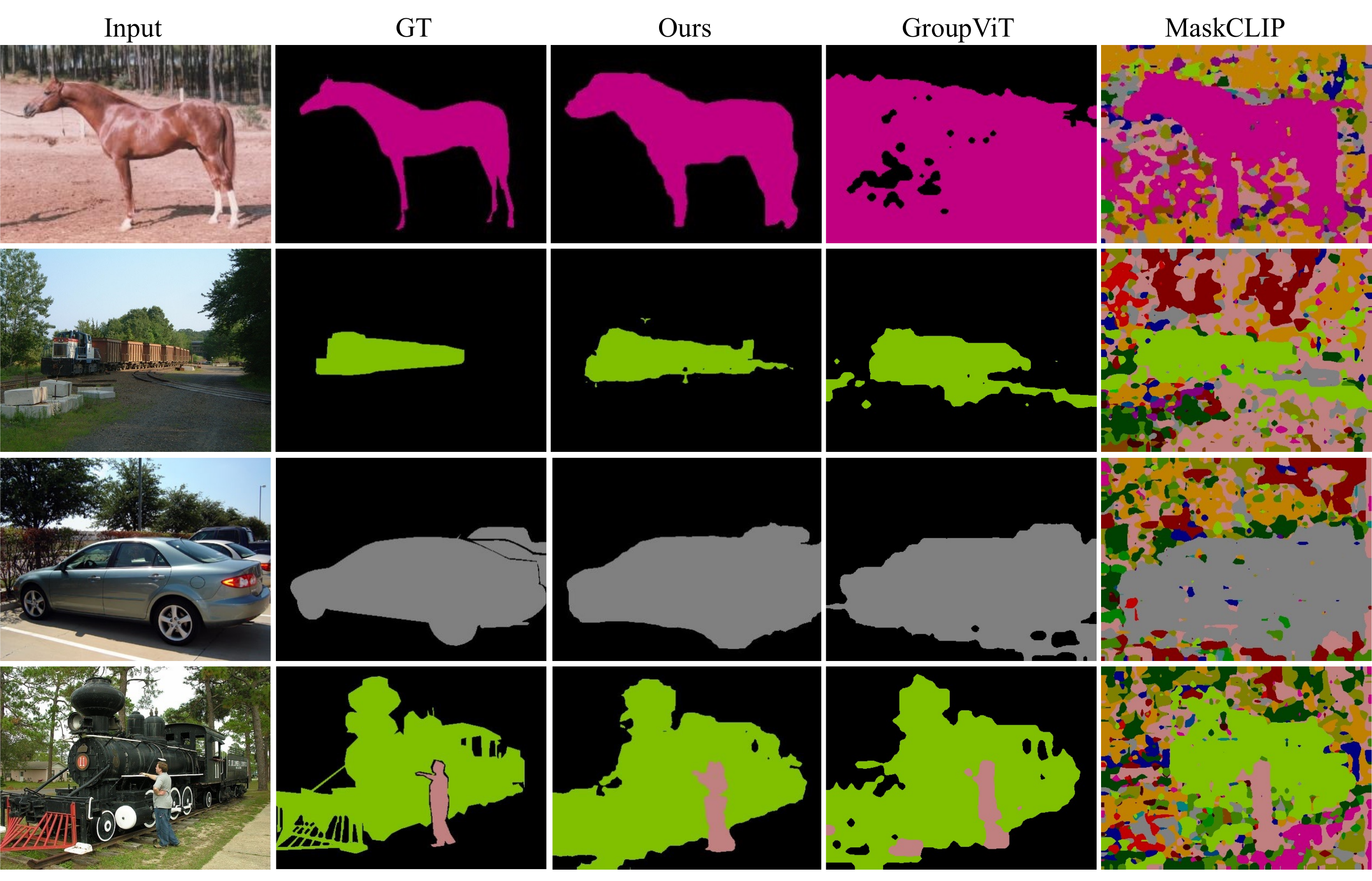}
\caption{\textbf{Comparsion of semantic segmentation on PASCAL VOC 2012.} Our method produces more consistent masks than GroupViT~\cite{xu2022groupvit} and MaskCLIP~\cite{zhou2022extract}, while containing less small spurious regions.}
   \label{fig:vis_com}
   \vspace{-1em}
\end{figure*}

\begin{figure*}[t!]
  \centering
   \includegraphics[width=0.95\linewidth]{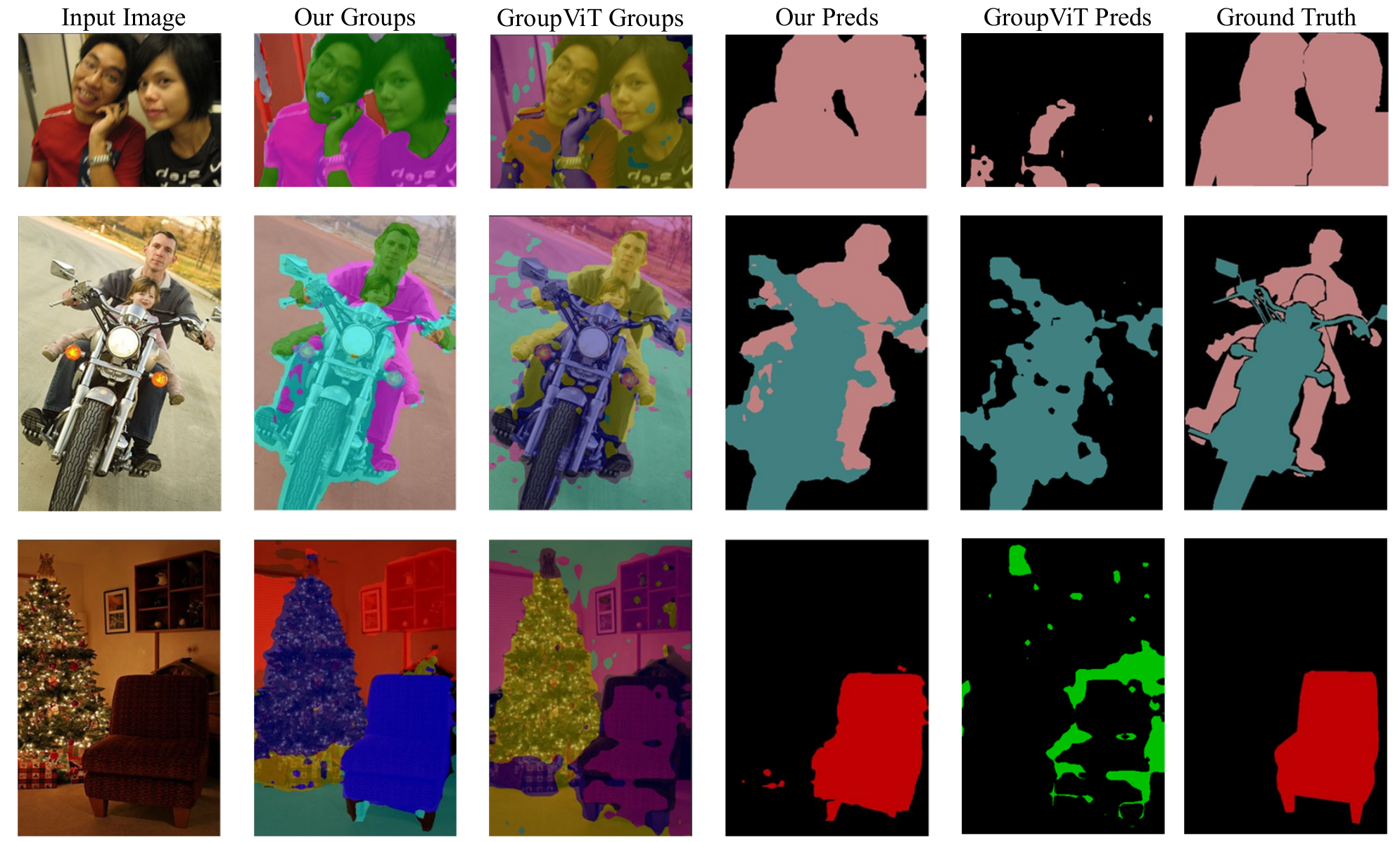}
   \caption{\textbf{Comparsion of grouped results on PASCAL VOC 2012.} The groups of our method are directly parsed by clustering the self-supervised features. The groups of GroupViT are generated from the attention map of the last grouping blocks. Both methods assign each grouped region to its closest label embedding to predict the final segmentation results.}
   \label{fig:qualf}
   \vspace{-1em}
\end{figure*}
\subsection{Qualitative Results}
\label{sec:qual}
We qualitatively compare our method with previous approaches in Fig.~\ref{fig:vis_com}.
As one can see, our method produces higher quality segmentation masks than others, \textit{i.e.}, fewer small spurious regions and more accurate boundaries.
In contrast, GroupViT~\cite{xu2022groupvit} generates masks with inconsistent regions, \eg, in row $4$th of Fig.~\ref{fig:vis_com}, small regions belonging to ``train'' are parsed as ``person''.
MaskCLIP~\cite{zhou2022extract} performs worse and produces severely noisy masks.

To analyze the advantage of the consistent grouping and region-level enhancement adaption, we further visualize the grouping and segmentation comparisons in Fig.~\ref{fig:qualf}.
%
%
As we mentioned in Sec.~\ref{sec:preliminary}, clustering with self-supervised features generates more compact regions with less noisy pixels.
For example, in the $3$rd image in Fig.~\ref{fig:qualf}, the foreground object ``sofa'' has consistent texture and shape. 
%
%
The segmentation results show that our region-level adaption is more advantageous than the image-sentence level alignment when recognizing the grouped regions:
When an entire object is divided into several local groups, our method correctly recognizes them as the parts of the object. For example, in the $1$st image, the faces and clothes of the ``person'' are grouped into different regions. Our method could correctly recognize all those groups as parts of a ``person''.
%
%

\subsection{Ablation Studies}
\label{sec:ablation}
\begin{table}[t]
    \centering
   \caption{\textbf{Masking strategy}.
   We compare three masking strategies described in Fig.~\ref{fig:mask}.
   Strategy (a) directly converts each grouped region into an foreground image, and encodes them with $M$ forward passes.
   Strategy (b) operates masks on patch tokens with one pass.
   Our (c) simultaneously queries all grouped regions from context-aware image features.
   }
   \begin{tabular}{c|cc}
      Masking Strategy & mIoU & Time Cost \\
      \shline
      Fig.~\ref{fig:mask} (a) &39.1 &59.1 ms \\
       Fig.~\ref{fig:mask} (b) &27.6 &11.5 ms\\
      Fig.~\ref{fig:mask} (c) &\textbf{54.8} &\textbf{11.5} ms\\    
   \end{tabular}
    \label{tab:ab_mask}
 \end{table}
\begin{table}[t]
   \centering
   \caption{\textbf{Fine-tuning parameters}.
    ``Entire Model'' means that fine-tuning all parameters of CLIP.
    ``Learnable Token'' denotes that introducing $M$ learnable tokens for each grouped region while freezing other parameters.
   }
   \begin{tabular}{l|c}
      Fine-tuning Parameters & mIoU \\
      \shline
      w/o Fine-tuning &54.8 \\
      Entire Model & 36.1 \\    
      Learnable Token &\textbf{59.2} \\
   \end{tabular}
    \label{tab:ab_finetune}
 \end{table}
\begin{table}[t]
   \centering
   \caption{\textbf{Assignment direction}.
   Assignment direction represents that how to match nouns and grouped regions in an image-sentence pair.
   ``Noun-to-region'' means that only assigning each noun to its closest region in image, while the direction of ``Region-to-noun'' is reverse.
   ``Bidirection'' uses both directions.
    Considering the asymmetric object categories in image-sentence pairs, ''Noun-to-region'' is adopted in our method. 
   }
   \begin{tabular}{l|c}
      Assignment Direction & mIoU\\
      \shline
      Noun-to-region &\textbf{59.2} \\
      Region-to-noun &52.5 \\
      Bidirection & 56.5 \\
   \end{tabular}
    \label{tab:ab_assign}
 \end{table}
In all ablation studies, our method is initialized with DINO-IN1M and CLIP-400M.
We evaluate its performance with mIoU on Pascal VOC validation set.
\paragraph{Masking strategy.}
We investigate three masking strategies from the perspective of efficiency and effectiveness in Tab.\ref{tab:ab_mask}.
All three models are not fine-tuned with noun-region contrastive loss.
Compared to the other two strategies, our masking strategy (c) greatly boosts the performance in semantic segmentation using context-aware image features.
Furthermore, all region-bound class tokens concurrently query their corresponding representations from image features, 
bringing $\sim 6$ times the improvement of inference speed than strategy (a).
%
%
\paragraph{Fine-tuning parameters.}
Our method aims to inherit the open-vocabulary recognition ability from CLIP, and then adapt it to region-level recognition.
%
%
We attempt to fine-tune different parameters during adaption in Tab.~\ref{tab:ab_finetune}.
Compared to the initial CLIP, fine-tuning the entire CLIP on CC12M causes extremely performance degradation on Pascal VOC, since the fine-tuned model may suffer from catastrophic forgetting~\cite{ilharco2022patching}.
%
In contrast, introducing several learnable tokens obtains a good trade-off between region-level adaption and open-vocabulary preservation, bringing a consistent $4.4\%$ mIoU gain.
\paragraph{Assignment direction.}
To verify the effectiveness of the one-way noun-region contrastive loss, we investigate three assignment directions between extracted nouns and grouped regions in Tab.~\ref{tab:ab_assign}.
As one can see, the ``noun-to-region'' direction outperforms the other two directions in a large margin, $7.0\%$ mIoU and $2.7\%$ mIoU respectively.
The experimental results are consistent with our observation in Fig.~\ref{fig:image_caption}: the object-grained information in a sentence is sparser than in its paired image, and adding ``region-to-noun'' assignment may lead to unnecessary alignment between mismatched noun-region pairs.

\section{Conclusion}
In this paper, we revealed that inconsistent grouped regions limit the upper bound performance of text-supervised semantic segmentation, and proposed to break it by exploiting spatially-consistent grouping of self-supervised vision models.
Considering part-like regions from self-supervised vision models, we effectively adapt an image-level CLIP to region-level recognition by two core designs.
Firstly, we present a context-ware masking strategy to simultaneously query representations of all grouped regions.
Secondly, we introduce a one-way noun-region contrastive loss for fine-grained alignment, while reducing mis-aligned noun-region pairs.
We conduct extensive experiments on popular benchmarks and demonstrate that our method achieves state-of-the-art results quantitatively and qualitatively.
Given the tremendous progress in vision-language pre-training, we believe that there will be substantial approaches reducing the gap between text-supervised and fully-supervised semantic segmentation.
%
%
%
%
%


% ...

{\small
\bibliographystyle{ieee_fullname}
\bibliography{egbib}

\begin{thebibliography}{10}\itemsep=-1pt

\bibitem{amir2021deep}
Shir Amir, Yossi Gandelsman, Shai Bagon, and Tali Dekel.
\newblock Deep vit features as dense visual descriptors.
\newblock {\em arXiv preprint arXiv:2112.05814}, 2(3):4, 2021.

\bibitem{bearman2016s}
Amy Bearman, Olga Russakovsky, Vittorio Ferrari, and Li Fei-Fei.
\newblock What’s the point: Semantic segmentation with point supervision.
\newblock In {\em European conference on computer vision}, pages 549--565,
  2016.

\bibitem{caesar2018coco}
Holger Caesar, Jasper Uijlings, and Vittorio Ferrari.
\newblock Coco-stuff: Thing and stuff classes in context.
\newblock In {\em Proceedings of the IEEE conference on computer vision and
  pattern recognition}, pages 1209--1218, 2018.

\bibitem{caron2021emerging}
Mathilde Caron, Hugo Touvron, Ishan Misra, Herv\'e J\'egou, Julien Mairal,
  Piotr Bojanowski, and Armand Joulin.
\newblock Emerging properties in self-supervised vision transformers.
\newblock In {\em Proceedings of the International Conference on Computer
  Vision (ICCV)}, 2021.

\bibitem{changpinyo2021conceptual}
Soravit Changpinyo, Piyush Sharma, Nan Ding, and Radu Soricut.
\newblock Conceptual 12m: Pushing web-scale image-text pre-training to
  recognize long-tail visual concepts.
\newblock In {\em Proceedings of the IEEE/CVF Conference on Computer Vision and
  Pattern Recognition}, pages 3558--3568, 2021.

\bibitem{chen2020simple}
Ting Chen, Simon Kornblith, Mohammad Norouzi, and Geoffrey Hinton.
\newblock A simple framework for contrastive learning of visual
  representations.
\newblock In {\em International conference on machine learning}, pages
  1597--1607, 2020.

\bibitem{deng2009imagenet}
Jia Deng, Wei Dong, Richard Socher, Li-Jia Li, Kai Li, and Li Fei-Fei.
\newblock Imagenet: A large-scale hierarchical image database.
\newblock In {\em 2009 IEEE conference on computer vision and pattern
  recognition}, pages 248--255, 2009.

\bibitem{dhanachandra2015image}
Nameirakpam Dhanachandra, Khumanthem Manglem, and Yambem~Jina Chanu.
\newblock Image segmentation using k-means clustering algorithm and subtractive
  clustering algorithm.
\newblock {\em Procedia Computer Science}, 54:764--771, 2015.

\bibitem{ding2022open}
Zheng Ding, Jieke Wang, and Zhuowen Tu.
\newblock Open-vocabulary panoptic segmentation with maskclip.
\newblock {\em arXiv preprint arXiv:2208.08984}, 2022.

\bibitem{everingham2010pascal}
Mark Everingham, Luc Van~Gool, Christopher~KI Williams, John Winn, and Andrew
  Zisserman.
\newblock The pascal visual object classes (voc) challenge.
\newblock {\em International journal of computer vision}, 88(2):303--338, 2010.

\bibitem{ghiasi2022scaling}
Golnaz Ghiasi, Xiuye Gu, Yin Cui, and Tsung-Yi Lin.
\newblock Scaling open-vocabulary image segmentation with image-level labels.
\newblock In {\em European Conference on Computer Vision}, pages 540--557,
  2022.

\bibitem{hamilton2022unsupervised}
Mark Hamilton, Zhoutong Zhang, Bharath Hariharan, Noah Snavely, and William~T
  Freeman.
\newblock Unsupervised semantic segmentation by distilling feature
  correspondences.
\newblock {\em arXiv preprint arXiv:2203.08414}, 2022.

\bibitem{he2022masked}
Kaiming He, Xinlei Chen, Saining Xie, Yanghao Li, Piotr Doll{\'a}r, and Ross
  Girshick.
\newblock Masked autoencoders are scalable vision learners.
\newblock In {\em Proceedings of the IEEE/CVF Conference on Computer Vision and
  Pattern Recognition}, pages 16000--16009, 2022.

\bibitem{he2020momentum}
Kaiming He, Haoqi Fan, Yuxin Wu, Saining Xie, and Ross Girshick.
\newblock Momentum contrast for unsupervised visual representation learning.
\newblock In {\em Proceedings of the IEEE/CVF conference on computer vision and
  pattern recognition}, pages 9729--9738, 2020.

\bibitem{ilharco2022patching}
Gabriel Ilharco, Mitchell Wortsman, Samir~Yitzhak Gadre, Shuran Song, Hannaneh
  Hajishirzi, Simon Kornblith, Ali Farhadi, and Ludwig Schmidt.
\newblock Patching open-vocabulary models by interpolating weights.
\newblock {\em arXiv preprint arXiv:2208.05592}, 2022.

\bibitem{jia2021scaling}
Chao Jia, Yinfei Yang, Ye Xia, Yi-Ting Chen, Zarana Parekh, Hieu Pham, Quoc Le,
  Yun-Hsuan Sung, Zhen Li, and Tom Duerig.
\newblock Scaling up visual and vision-language representation learning with
  noisy text supervision.
\newblock In {\em International Conference on Machine Learning}, pages
  4904--4916, 2021.

\bibitem{jia2022visual}
Menglin Jia, Luming Tang, Bor-Chun Chen, Claire Cardie, Serge Belongie, Bharath
  Hariharan, and Ser-Nam Lim.
\newblock Visual prompt tuning.
\newblock {\em arXiv preprint arXiv:2203.12119}, 2022.

\bibitem{liang2022open}
Feng Liang, Bichen Wu, Xiaoliang Dai, Kunpeng Li, Yinan Zhao, Hang Zhang,
  Peizhao Zhang, Peter Vajda, and Diana Marculescu.
\newblock Open-vocabulary semantic segmentation with mask-adapted clip.
\newblock {\em arXiv preprint arXiv:2210.04150}, 2022.

\bibitem{lin2014microsoft}
Tsung-Yi Lin, Michael Maire, Serge Belongie, James Hays, Pietro Perona, Deva
  Ramanan, Piotr Doll{\'a}r, and C~Lawrence Zitnick.
\newblock Microsoft coco: Common objects in context.
\newblock In {\em European conference on computer vision}, pages 740--755,
  2014.

\bibitem{long2015fully}
Jonathan Long, Evan Shelhamer, and Trevor Darrell.
\newblock Fully convolutional networks for semantic segmentation.
\newblock In {\em Proceedings of the IEEE conference on computer vision and
  pattern recognition}, pages 3431--3440, 2015.

\bibitem{loshchilov2017decoupled}
Ilya Loshchilov and Frank Hutter.
\newblock Decoupled weight decay regularization.
\newblock {\em arXiv preprint arXiv:1711.05101}, 2017.

\bibitem{minaee2021image}
Shervin Minaee, Yuri~Y Boykov, Fatih Porikli, Antonio~J Plaza, Nasser
  Kehtarnavaz, and Demetri Terzopoulos.
\newblock Image segmentation using deep learning: A survey.
\newblock {\em IEEE transactions on pattern analysis and machine intelligence},
  2021.

\bibitem{minderer2022simple}
Matthias Minderer, Alexey Gritsenko, Austin Stone, Maxim Neumann, Dirk
  Weissenborn, Alexey Dosovitskiy, Aravindh Mahendran, Anurag Arnab, Mostafa
  Dehghani, Zhuoran Shen, et~al.
\newblock Simple open-vocabulary object detection with vision transformers.
\newblock {\em arXiv preprint arXiv:2205.06230}, 2022.

\bibitem{mottaghi2014role}
Roozbeh Mottaghi, Xianjie Chen, Xiaobai Liu, Nam-Gyu Cho, Seong-Whan Lee, Sanja
  Fidler, Raquel Urtasun, and Alan Yuille.
\newblock The role of context for object detection and semantic segmentation in
  the wild.
\newblock In {\em Proceedings of the IEEE conference on computer vision and
  pattern recognition}, pages 891--898, 2014.

\bibitem{radford2021learning}
Alec Radford, Jong~Wook Kim, Chris Hallacy, Aditya Ramesh, Gabriel Goh,
  Sandhini Agarwal, Girish Sastry, Amanda Askell, Pamela Mishkin, Jack Clark,
  et~al.
\newblock Learning transferable visual models from natural language
  supervision.
\newblock In {\em International Conference on Machine Learning}, pages
  8748--8763, 2021.

\bibitem{ranasinghe2022perceptual}
Kanchana Ranasinghe, Brandon McKinzie, Sachin Ravi, Yinfei Yang, Alexander
  Toshev, and Jonathon Shlens.
\newblock Perceptual grouping in vision-language models.
\newblock {\em arXiv preprint arXiv:2210.09996}, 2022.

\bibitem{thomee2016yfcc100m}
Bart Thomee, David~A Shamma, Gerald Friedland, Benjamin Elizalde, Karl Ni,
  Douglas Poland, Damian Borth, and Li-Jia Li.
\newblock Yfcc100m: The new data in multimedia research.
\newblock {\em Communications of the ACM}, 59(2):64--73, 2016.

\bibitem{van2022discovering}
Wouter Van~Gansbeke, Simon Vandenhende, and Luc Van~Gool.
\newblock Discovering object masks with transformers for unsupervised semantic
  segmentation.
\newblock {\em arXiv preprint arXiv:2206.06363}, 2022.

\bibitem{wang2022tokencut}
Yangtao Wang, Xi Shen, Yuan Yuan, Yuming Du, Maomao Li, Shell~Xu Hu, James~L
  Crowley, and Dominique Vaufreydaz.
\newblock Tokencut: Segmenting objects in images and videos with
  self-supervised transformer and normalized cut.
\newblock {\em arXiv preprint arXiv:2209.00383}, 2022.

\bibitem{xu2022groupvit}
Jiarui Xu, Shalini De~Mello, Sifei Liu, Wonmin Byeon, Thomas Breuel, Jan Kautz,
  and Xiaolong Wang.
\newblock Groupvit: Semantic segmentation emerges from text supervision.
\newblock In {\em Proceedings of the IEEE/CVF Conference on Computer Vision and
  Pattern Recognition}, pages 18134--18144, 2022.

\bibitem{yao2021filip}
Lewei Yao, Runhui Huang, Lu Hou, Guansong Lu, Minzhe Niu, Hang Xu, Xiaodan
  Liang, Zhenguo Li, Xin Jiang, and Chunjing Xu.
\newblock Filip: Fine-grained interactive language-image pre-training.
\newblock {\em arXiv preprint arXiv:2111.07783}, 2021.

\bibitem{zhai2022lit}
Xiaohua Zhai, Xiao Wang, Basil Mustafa, Andreas Steiner, Daniel Keysers,
  Alexander Kolesnikov, and Lucas Beyer.
\newblock Lit: Zero-shot transfer with locked-image text tuning.
\newblock In {\em Proceedings of the IEEE/CVF Conference on Computer Vision and
  Pattern Recognition}, pages 18123--18133, 2022.

\bibitem{zhong2022regionclip}
Yiwu Zhong, Jianwei Yang, Pengchuan Zhang, Chunyuan Li, Noel Codella,
  Liunian~Harold Li, Luowei Zhou, Xiyang Dai, Lu Yuan, Yin Li, et~al.
\newblock Regionclip: Region-based language-image pretraining.
\newblock In {\em Proceedings of the IEEE/CVF Conference on Computer Vision and
  Pattern Recognition}, pages 16793--16803, 2022.

\bibitem{zhou2019semantic}
Bolei Zhou, Hang Zhao, Xavier Puig, Tete Xiao, Sanja Fidler, Adela Barriuso,
  and Antonio Torralba.
\newblock Semantic understanding of scenes through the ade20k dataset.
\newblock {\em International Journal of Computer Vision}, 127(3):302--321,
  2019.

\bibitem{zhou2022extract}
Chong Zhou, Chen~Change Loy, and Bo Dai.
\newblock Extract free dense labels from clip.
\newblock In {\em European Conference on Computer Vision}, pages 696--712,
  2022.

\end{thebibliography}
}

\clearpage

\section*{A. Pre-training Details of Small CLIPs}
We pre-train a CLIP-12M on Conceptual Caption 12M, and CLIP-27M on Conceptual Caption 12M +  Yahoo Flickr Creative Commons 15M following LiT~\cite{zhai2022lit}.
For image side, we choose DINO~\cite{caron2021emerging} with ViT-B/16 architecture as our image encoder and freeze it during training.
For text side, we follow the architecture of CLIP~\cite{radford2021learning} text encoder but randomly initialize it.
During training, we use AdamW optimizer with learning rate $0.00005$ and weight decay $0.05$.
We train small CLIPs on their corresponding datasets for $10$ epochs with batch size $4,096$.
Thanks to fast convergence of LiT, the training of our small CLIPs is done within $12$ hours on $8$ V100 GPUs.
We report the zero-shot accuracy on ImageNet classification.

\begin{table}[htbp]
   \centering
   \caption{\textbf{Accuracy comparisons on ImageNet classification}.
   }
   \begin{tabular}{l|c}
      Method & Accuracy\\
      \shline
      GroupViT~\cite{xu2022groupvit} &42.9 \\
      CLIPpy~\cite{ranasinghe2022perceptual} &45.3 \\
      CLIP-12M(ours) & \underline{46.0} \\
      CLIP-27M(ours) & \textbf{47.8} \\
   \end{tabular}
    \label{tab:sup_class}
 \end{table}

\section*{B. More Visualizations}
\begin{figure*}[htbp]
  \centering
   \includegraphics[width=0.7\linewidth]{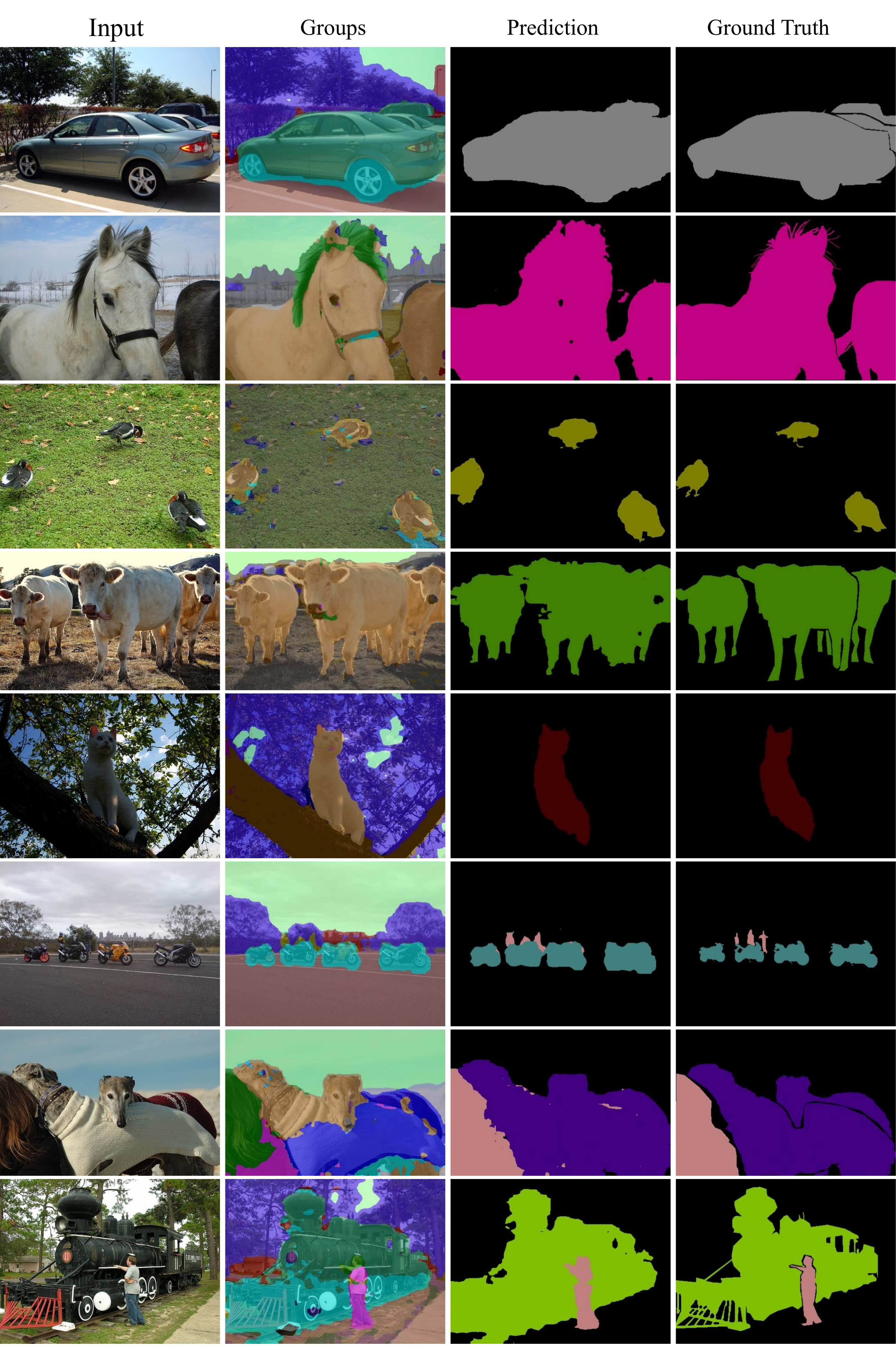}
   \caption{\textbf{More visualizations of our method on Pascal VOC.}
   From left to right, each column shows input images, grouped regions, prediction results, and ground truth respectively.
   }
   \label{fig:sup_voc}
\end{figure*}

\begin{figure*}[htbp]
  \centering
   \includegraphics[width=0.7\linewidth]{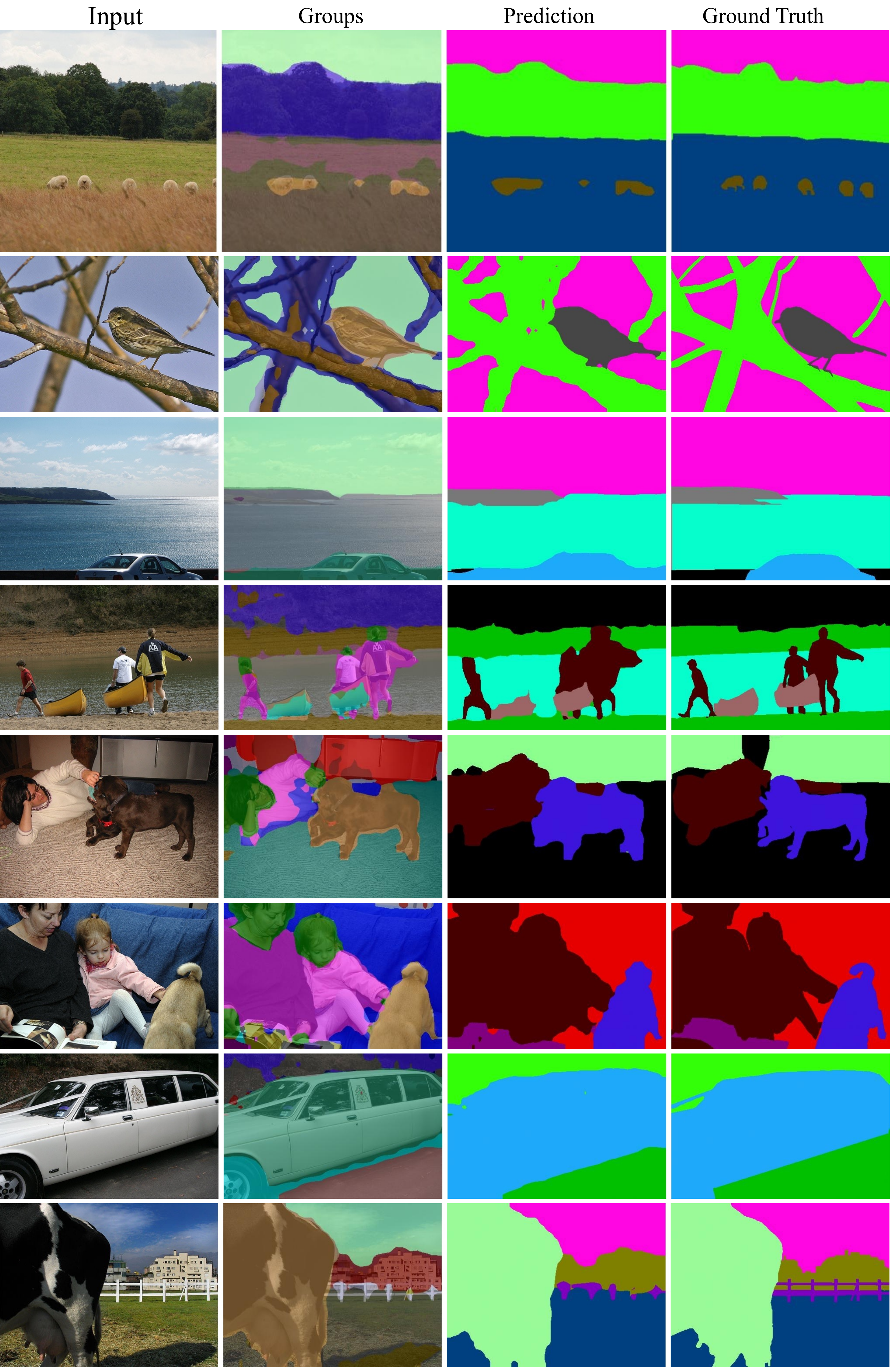}
   \caption{\textbf{Visualizations of our method on Pascal Context.}
   From left to right, each column shows input images, grouped regions, prediction results, and ground truth respectively.
   }
   \label{fig:sup_context}
\end{figure*}

\begin{figure*}[htbp]
  \centering
   \includegraphics[width=0.7\linewidth]{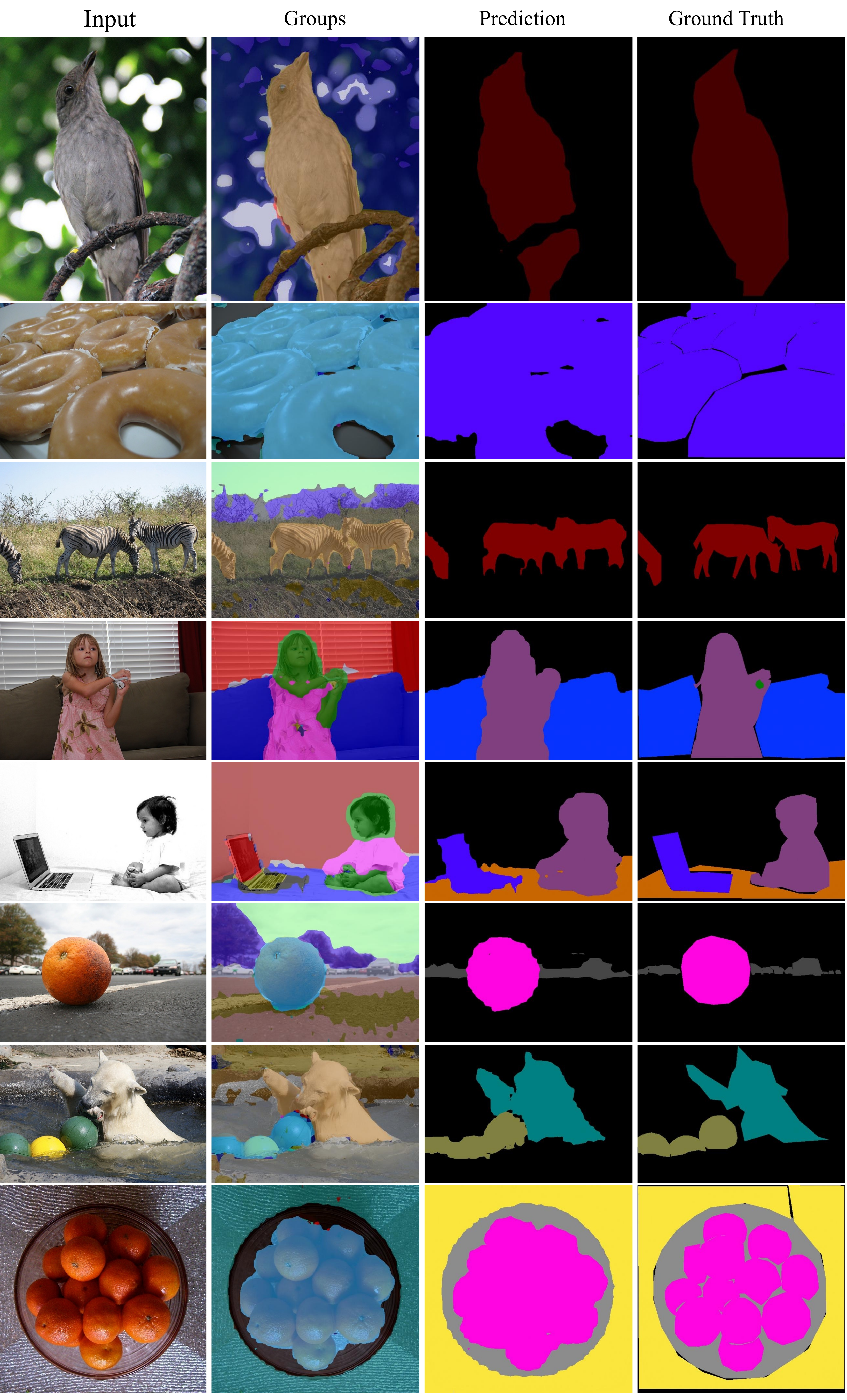}
   \caption{\textbf{Visualizations of our method on COCO.}
   From left to right, each column shows input images, grouped regions, prediction results, and ground truth respectively.
   }
   \label{fig:sup_coco}
\end{figure*}

We visualize additional predictions of our method on Pascal VOC in Fig.~\ref{fig:sup_voc}.
Meanwhile, we also show more visualizations of our method on Pascal Context in Fig.~\ref{fig:sup_context} and COCO in Fig.~\ref{fig:sup_coco}.
As one can see, our method not only groups input images into high-quality regions, but also recognizes these regions correctly.
Moreover, as shown in row $1$ and $2$ of Fig.~\ref{fig:sup_coco}, our predictions even achieves better results than the ground truth, \eg, ``branch'' in ``bird'' mask and ``table'' in ``donut'' mask.
%
%
%
%

%

\end{document}